\PassOptionsToPackage{unicode}{hyperref}
\PassOptionsToPackage{hyphens}{url}
\documentclass[
  twocolumn]{article}
\usepackage{amsmath,amssymb}
\usepackage{lmodern}
\usepackage{iftex}
\ifPDFTeX
  \usepackage[T1]{fontenc}
  \usepackage[utf8]{inputenc}
  \usepackage{textcomp} 
\else 
  \usepackage{unicode-math}
  \defaultfontfeatures{Scale=MatchLowercase}
  \defaultfontfeatures[\rmfamily]{Ligatures=TeX,Scale=1}
\fi
\IfFileExists{upquote.sty}{\usepackage{upquote}}{}
\IfFileExists{microtype.sty}{
  \usepackage[]{microtype}
  \UseMicrotypeSet[protrusion]{basicmath} 
}{}
\makeatletter
\@ifundefined{KOMAClassName}{
  \IfFileExists{parskip.sty}{%
    \usepackage{parskip}
  }{
    \setlength{\parindent}{0pt}
    \setlength{\parskip}{6pt plus 2pt minus 1pt}}
}{
  \KOMAoptions{parskip=half}}
\makeatother
\usepackage{xcolor}
\usepackage[margin=1in]{geometry}
\usepackage{graphicx}
\makeatletter
\def\maxwidth{\ifdim\Gin@nat@width>\linewidth\linewidth\else\Gin@nat@width\fi}
\def\maxheight{\ifdim\Gin@nat@height>\textheight\textheight\else\Gin@nat@height\fi}
\makeatother
\setkeys{Gin}{width=\maxwidth,height=\maxheight,keepaspectratio}
\makeatletter
\def\fps@figure{htbp}
\makeatother
\ifLuaTeX
  \usepackage{selnolig}  
\fi
\IfFileExists{bookmark.sty}{\usepackage{bookmark}}{\usepackage{hyperref}}
\IfFileExists{xurl.sty}{\usepackage{xurl}}{} 
\hypersetup{
  pdftitle={Longitudinal Sentiment Analyses for Radicalization Research},
  pdfauthor={a Discussion Paper by Dennis Klinkhammer},
  hidelinks,
  pdfcreator={LaTeX via pandoc}}

\title{Longitudinal Sentiment Analyses for Radicalization Research}
\usepackage{etoolbox}
\makeatletter
\providecommand{\subtitle}[1]{
  \apptocmd{\@title}{\par {\large #1 \par}}{}{}
}
\makeatother
\subtitle{Intertemporal Dynamics on Social Media Platforms and their
Implications}
\author{a Discussion Paper by Dennis Klinkhammer}
\date{}

\begin{document}
\maketitle

\hypertarget{abstract}{%
\paragraph{Abstract}\label{abstract}}

\emph{Sentiment analysis is a sub-discipline in the field of natural
language processing and computational linguistics and can be used for
automated or semi-automated analyses of qualitative data. The primary
context of application is to recognize an expressed attitude as positive
or negative as it can be contained in qualitative data, such as comments
on social media platforms. However, a cross-sectional perspective
regarding sentiments within social media comments has proven to be error
prone when it comes to the detection of radicalization, extremism and
hate speech. Since these phenomena are processes over time, there seems
to be an increasing demand for longitudinal perspectives in the context
of radicalization research. This discussion paper demonstrates how
longitudinal sentiment analyses can depict intertemporal dynamics on
social media platforms, what challenges are inherent and how further
research could benefit from a longitudinal perspective. Furthermore and
since tools for sentiment analyses shall simplify and accelerate the
analytical process regarding qualitative data at acceptable inter-rater
reliability, their applicability in the context of radicalization
research will be examined regarding the Tweets collected on January 6th
2021, the day of the storming of the U.S. Capitol in Washington.
Therefore, a total of 49,350 Tweets will be analyzed evenly distributed
within three different sequences: before, during and after the U.S.
Capitol in Washington was stormed. These sequences highlight the
intertemporal dynamics within comments on social media platforms as well
as the possible benefits of a longitudinal perspective when using
conditional means and conditional variances. Limitations regarding the
identification of supporters of such events and associated hate speech
as well as common application errors will be demonstrated as well. As a
result, only under certain conditions a longitudinal sentiment analysis
can increase the accuracy of evidence based predictions in the context
of radicalization research.}

\hypertarget{keywords}{%
\paragraph{Keywords}\label{keywords}}

Sentiment Analysis, Natural Language Processing, Social Media,
Radicalization Research

\hypertarget{i-introduction}{%
\subsection{(I) Introduction}\label{i-introduction}}

Sentiment analyses can be used to capture the sentiment in qualitative
data, such as comments on social media platforms. Within comments on
social media platforms the polarity can be classified word by word as
positive or negative and in some cases as neutral via basic sentiment
analysis. In addition, different types of emotional states can be
classified via advanced sentiment analysis by using the NRC Word-Emotion
Association Lexicon, the first word-emotion lexicon with eight basic
emotional states (Mohammad 2020). Despite positive and negative
sentiments, it is possible to classify anger, fear, anticipation, trust,
surprise, sadness, joy, and disgust as well.

Furthermore, sentiment analyses have a sufficient
inter-rater-reliability with less time requirements. The
inter-rater-reliability is a degree of agreement among independent
observers who rate, code, or assess the same phenomenon within
qualitative data. While scientist usually achieve an
inter-rater-reliability up to 80\%, sentiment analyses can achieve up to
70\%. This appears to be an acceptable value, because even if different
types of sentiment analyses would agree up to 100\%, research indicates
that scientists would still disagree by 20\% (Ogneva 2010). Therefore,
sentiment analyses can be found in a broad application context. However,
the question regarding their accuracy in the context of radicalization
research has not yet been adequately answered, but sentiment analyses
are nevertheless increasingly used regarding related phenomena on social
media platforms (Torregrosa et al.~2022).

Especially social media platforms like Twitter offer several users a
low-threshold opportunity to exchange opinions and experiences. For
example, these opinions and experiences can affect various areas of
society, such as political and economical ones. Research indicates that,
for example, comments on social media platforms can be used to capture
social issues like radicalization and extremism (Tanoli et al.~2022),
sexuality (Wood et al.~2017), side effects of medication and drugs
(Korkontzelos et al.~2016) as well as for the reflection of the offline
political landscape (Tumasjan et al.~2010). A well known use-case is the
political campaign of former U.S. president Barack Obama, who used
sentiment analysis back in 2012. There are many possible use-cases, but
also numerous challenges in the application of sentiment analyses.
Accordingly, research in this area continues (Klinkhammer 2022; Hamborg
\& Donnay 2021).

\hypertarget{ii-theoretical-background}{%
\subsection{(II) Theoretical
Background}\label{ii-theoretical-background}}

Social media platforms like Twitter have demonstrated a continuous
increase of active users over the most recent years (Pereira-Kohatsu et
al.~2019). An average of 500 million Tweets per day combined with a low
threshold regarding the participation leads to a high diversity of
opinions (Koehler 2015). As a result, Twitter is not to be interpreted
as one singular social network, but as several social sub-networks, who
enable users to exchange information with each other.

Some of these sub-networks are so-called echo chambers (Bright 2017).
Echo chambers can arise through an accumulation of thematically related
Tweets, replies, likes and followers. Since Twitter as social media
platform allows its users to switch quickly and uncomplicated from one
social sub-network into another (Prior 2005), it is to be assumed that
echo chamber are most likely to arise and unfold their dynamics. As a
result, users usually participate within echo chambers, which correspond
with their own opinion and the so called echo arises. Within an echo
chamber the own opinion can be confirmed and this confirmation bias can
lead to distortions regarding the perception of social phenomena outside
a social media platform like Twitter (Cinelli et al.~2021; Jacobs \&
Spierings 2018).

It has already be confirmed that these confirmation biases within echo
chambers with a political agenda can lead to a gradual accumulation from
radical to extreme to anti-constitutional opinions (O'Hara \& Stevens
2015). According to Neumann (2013), extremism is a context-specific term
and must be compared and adapted to the accepted socio-political
realities of the observed society. Extremism emerges from the process of
radicalization and can be divided into cognitive and violent extremism
(Neumann et al.~2018): It can also be stated that extremism is
characterized by a willingness to act in order to endanger life, freedom
and rights of others.

Furthermore, these echo chambers enable users to perform a continuing
defamation of dissenters and in some cases these defamation strategies
follow the aim of political influence (Glaser \& Pfeiffer 2017). This
negative communication is called hate speech and aims at the exclusion
of single persons or groups of persons, because of their ethnicity,
sexual orientation, gender identity, disability, religion or political
views (Pereira-Kohatsu et al.~2019; Warner \& Hirschberg 2012).
According to Kay (2011) and Sunstein (2006), extremist networks show a
low tolerance towards individuals and groups who think differently and
are generally less cosmopolitan.

As a result of these echo chambers, hate speech as well as radicalizing
elements show an increasing number on social media platforms
(Reichelmann et al.~2020; Barberá et al.~2015). Therefore, Twitter is
often accused of being a platform for polarizing, racist, antisemitic or
anti-constitutional content (Awan 2017; Gerstenfeld et al.~2003). This
content is usually also freely accessible to children and young people
(Machackova et al.~2020). It could now be hypothesized that certain
messages will get more attention, even if only a small minority of
activists uses the echo chambers accordingly.

However, such social media elements could also be used to investigate
communications patterns inside social networks and social sub-networks,
in order to focus the role of individual users and what influence the
content of their comments and actions might have on the underlying
structures of a social network (Klinkhammer 2020). Therefore, social
media platforms could also become a sensor of the real world and provide
important information for criminological investigations and predictions
(Scanlon \& Gerber 2015; Sui et al.~2014). Corresponding research papers
have been published recently (Hamachers et al.~2020) and five studies
represent the scientific efforts regarding the identification of hate
speech and extremism on Twitter (Charitidis et al.~2020; Mandl et
al.~2019; Wiegand et al.~2018; Bretschneider \& Peters 2017; Ross et
al.~2017).

Some of these research papers refer to mathematical and statistical
methods in order to identify hate speech as well as radicalizing
elements. Regression models and classification models are most commonly
used in machine learning based approaches and a few approaches are based
on deep learning via logistic regression models as basis for simple
neural networks (Schmidt \& Wiegand 2017), whereas more sophisticated
approaches make use of convolutional neural networks as well as
sentiment analyses (Hamachers et al.~2020).

While methodologically it is feasible to count the number of hate speech
and radicalizing elements and, for example, to study the impact of
anti-hate laws regarding social media platforms by using semi-automated
and merely descriptive approaches (Wienigk \& Klinkhammer 2021), the
process of automated identification without human supervision has proven
to be error-prone. For example, mean values and variance values as used
as reference values in many of these approaches only lead to a correct
identification in the short term (Klinkhammer 2020). However, the same
approaches can lead to false positive or false negative results when
conducted again at a later point in time. The decisive factor could be
the intertemporal dynamics on social media platforms, which also could
impact approaches based upon sentiment analyses (Grogan 2020).
Accordingly, in respect to the changing size and topics of echo chambers
over time and considering that radicalization is a process, a
longitudinal perspective seems recommended (Greipl et al.~2022). As an
intermediate result, two hypotheses can be derived for this discussion
paper:

\begin{enumerate}
\def\labelenumi{(\arabic{enumi})}
\item
  Patterns of communication on social media platforms can be subject to
  intertemporal dynamics.
\item
  Phenomena in the context of radicalization research can be
  superimposed by intertemporal dynamics.
\end{enumerate}

Therefore, it seems necessary to consider these hypotheses, not only to
perform a longitudinal sentiment analyses, but also in order to
implement the outcomes of sentiment analyses as independent variables
for evidence based predictions in the context of radicalization
research.

\hypertarget{iii-software-requirements}{%
\subsection{(III) Software
Requirements}\label{iii-software-requirements}}

This discussion paper bases upon previously published working papers and
tutorials on arXiv: \emph{Analysing Social Media Network Data with R:
Semi-Automated Screening of Users, Comments and Communication Patterns}
(Klinkhammer 2020) and \emph{Sentiment Analysis with R: Natural Language
Processing for Semi-Automated Assessments of Qualitative Data}
(Klinkhammer 2022). Both are based on R, an object based programming
language. Therefore, datasets, variables, cases, values as well as
functions can be applied as a combination of objects. This longitudinal
approach for sentiment analyses requires six additional packages in
order to expand the range of basic R functions, which are listed here in
order to make the requirements and the process of data pre-processing
and the subsequent analysis more transparent:

\begin{enumerate}
\def\labelenumi{(\arabic{enumi})}
\item
  Since the analysis of comments on social media platforms requires a
  focus on every single element that is to be analysed, it is necessary
  to break down the underlying data structure into manageable little
  pieces. A package that is specifically designed to do so is called
  \emph{dplyr}. It can split, apply and combine qualitative data for
  further analytical steps (Wickham 2022).
\item
  The second package is called \emph{stringr}. Since qualitative data,
  like social media comments, is represented by character variables in
  R, a package that can process and - if necessary - manipulate
  individual characters within the strings of a character variable is
  required (Wickham 2019); A string is marked either by single quote
  signs or double quote signs.
\item
  Another necessary package is called \emph{textdata}. It contains
  several words as references and sentiment libraries, such as the NRC
  Word-Emotion Association Lexicon (Mohammad 2020), as well as the Bing
  Sentiment Lexicon (Hu \& Liu 2004).
\item
  Sentiment analysis is a text mining technique and the package
  \emph{tidytext} is required in order to convert conventional
  qualitative data into tidy formats, such as single words without
  punctuation or spaces (De Queiroz et al.~2022). This allows scientists
  to focus on paragraphs or otherwise separated content word by word. As
  a result, the tidy text format lists and counts all words individually
  and assigns them a number according to the original Tweet.
\item
  The package \emph{tidytext} provides a connection between most
  commonly used packages like \emph{dplyr} and \emph{ggplot2} by using
  their basic formulas and commands. The latter is responsible for the
  detailed visualisation of sentiments and other types of results, based
  on ``The Grammer of Graphics'' (Wickham et al.~2022). In particular,
  defining the details of a visualisation enables scientists to create
  informative as well as attractive plots.
\item
  Finally, the package \emph{gridExtra} enables scientists to arrange
  multiple visualizations at once and to create dashboards for an
  intuitive display of relevant information (Auguie \& Antonov 2017).
\end{enumerate}

The previously mentioned working papers and tutorials provide a step by
step guide through the process of data pre-processing as well as the
analysis. Furthermore, the process of collecting data regarding social
media networks and comments is demonstrated as well.

\hypertarget{iv-dataset-and-data-pre-processing}{%
\subsection{(IV) Dataset and Data
Pre-Processing}\label{iv-dataset-and-data-pre-processing}}

Data pre-processing is used to check datasets for irrelevant and
redundant information present or noisy and unreliable data. In a first
step, 49.350 Tweets from January 6th 2021, the day the U.S. Capitol in
Washington was stormed, have been transformed into a tibble. Tibbles are
plain and simple datasets that can be processed by R. Furthermore,
sentiment analyses require a tidy data set. In a tidy data set, each
word within a Tweet is separated without losing connection to related
words in the same Tweet. As a result, the tidy dataset can be described
as a listing of all words within different Tweets. This allows the
sentiment analysis to be performed word by word for each Tweet. Since
non-essential words, such as \emph{and} or \emph{is}, may be included in
this list, they need to be eliminated. In addition, custom words and
duplicates can also be excluded from this object.

In a final step of data pre-processing, the dataset (49.350 Tweets) will
be splitted into three evenly distributed sequences: Before (16.450
Tweets), during (16.450 Tweets) and after (16.450 Tweets) the storming
of the U.S. Capitol in Washington. This subdivision is necessary because
each point in time is marked by different contextual conditions which
might have impact on the underlying dynamics of Tweets and thus for
intertemporal sentiment analyses. For example, if there is a strict
social norm about a topic at a given point in time, one might expect
smaller variance than when there are weak norms about a topic. However,
social norms can vary over time and in relation to acute events. Or one
might observe that people who highly identify with a group have a
smaller variance than those who are weakly identified with that group.
Again, it must be taken into account, that the identification with
groups might be subject to temporal fluctuations as well.

\hypertarget{v-methodological-approach}{%
\subsection{(V) Methodological
Approach}\label{v-methodological-approach}}

In order to highlight the dynamics regarding the appearance of
sentiments and emotional states over time, a smoothed slope will be
used. A smoothed slope does not represent actual values, e.g.~mean
values, but expected values based upon statistical modelling,
e.g.~conditional mean values. The conditional expectation can be either
a random variable, noted as \(E[X|Y]\), or a function, noted as
\(E[X|Y=y]\), where \(Y\) respectively \(Y=y\) represent the conditions.
If a random variable can take on only a finite number of values, the
conditions are that the variable can only take on a subset of those
values which the slope of the function would depict accordingly. One of
the advantages of expected values is their robustness to statistical
outliers. Readers interested in the history of statistics will recognize
parallels to the underlying work of Pierre-Simon Laplace, which was
formalized by Andrey Kolmogorov using the Radon-Nikodym theorem (Feller
1991).

Applying conditional expectations for longitudinal sentiment analyses is
necessary, because the slope of a function based on actual values could
be irratic due to sporadic occuring values. Nevertheless, only few
longitudinal sentiment analyses consider conditional values (Jacobs \&
Spierings 2018), although this method was suitable to depict the
dynamics of Tweets regarding politicians from populist parties.
Considering the large amount of Tweets (here: 49.350) that are to be
analyzed, sporadic occuring values are to be expected and will be
compensated by applying conditional expectations. Conditional
expectations will be calculated for each sequence of Tweets (here:
16.450) to establish intertemporal contextuality. It is to be assumed
that applying more small-stepped sequences could increase the accuracy
of the analysis further. The accuracy refers to the identification of
individual data points that could be relevant in the context of
radicalization research, e.g.~radical comments regarding the storming of
the U.S. Capitol in Washington. The analysis will highlight that
conditional means and conditional variances are more suitable for that
task than mean values and variance values.

In the actual analysis, not only the number of used sentiments over time
is to be analyzed, but also their summative score. The first analytical
step focuses the conditional mean values and conditional variance values
regarding the number of used sentiments over time, for which the NRC
Word-Emotion Association Lexicon is a suitable basis (Mohammad 2020).
The NRC Word-Emotion Association Lexicon differentiates between positive
and negative sentiments as well as eight emotional states: Anger, fear,
anticipation, trust, surprise, sadness, joy, and disgust. Since a Tweet
can consist of zero up to several sentiments, the expected values must
be greater than or equal to zero within each sequence of Tweets. Another
way of analysing the intertemporal use of sentiments and emotional
states is based upon summative scores. A sentiment score results from
the sum of positive (\emph{1}) and negative (\emph{-1}) values in
respect to the underlying sentiment. Since the NRC Word-Emotion
Association Lexicon from the first analytical step differentiates
between positive and negative sentiments as well as eight emotional
states, a validating lexicon can be added that differentiates only
between positive and negative sentiments: The Bing Sentiment Lexicon (Hu
\& Liu 2004). For each Tweet with sentiments the score of these
sentiments is assigned as sum of all positive (\emph{1}) and negative
(\emph{-1}) sentiments and the conditional mean values and conditional
variance values will be plotted as smoothed slope again. Thus, the
number of used sentiments over time is to be analyzed via smoothed
slopes and based upon the NRC Word-Emotion Lexicon as well as their
summative scores based upon the Bing Sentiment Lexicon, each applied on
all three sequences of the dataset in order to frame the intertemporal
dynamics. The possibility of a specific identification of individual
comments on social media platforms will be demonstrated by comparing the
values of means and variances within each sequence with their expected
values of the smoothed slope.

\hypertarget{vi-intertemporal-use-of-sentiments}{%
\subsection{(VI) Intertemporal Use of
Sentiments}\label{vi-intertemporal-use-of-sentiments}}

Focusing the intertemporal use of sentiments is supposed to reveal the
intertemporal dynamics within Tweets. Each sequence will compare the
mean values (red line) against the conditional mean values (blue slope).
As can be seen, the intertemporal dynamics of each sequence would not be
represented by one mean value for each sequence, whereas conditional
means vary clearly below and above average. Furthermore, two important
points for such analyzes are revealed: Intertemporal dynamics, as can be
seen by focusing the below and above average fluctuations, tend to be
small-stepped. For example, the mean values between the three sequences
vary from 1.56 to 1.70 and finally 1.50. In the present data, the second
sequence stands for the storming of the U.S. Capitol in Washington. Such
an event is reflected in the intertemporal dynamics by an increase in
the use of sentiments. The conditional means fluctuate a little bit more
and run below and above the mean values, covering a below average use of
sentiments as well as an above average use of sentiments. Corresponding
analyzes must therefore be very sensitive in order to cover these
fluctuations. Based on these sequences, it can now be examined which
topic is depicted on Twitter with which intertemporal use of sentiments.
Finally, individual Tweets and their positioning along the smoothed
slope can be identified in order to specify the patterns of
communication.

The first sequence is determined by two general topics: The necessity of
measurements regarding the COVID-19 pandemic as well as the elections
won by the Democratic Party in the federal state of Georgia. The lowest
values of the slope inidcate COVID-19 as dominant topic, whereas the
highest values represent the point in time the results of the elections
in Georgia and their confirmation were publicly announced. It becomes
clear that the ongoing COVID-19 pandemic as topic is characterized by a
lower use of sentiments on Twitter than the elections in Georgia. An
example for that specific event and the associated intertemporal
dynamics is a Tweet that can be located at the local maximum of the
slope ({[}a{]} Tweet 8.022): \emph{``jon ossoff will stand for all of
georgia in the fight for healthcare jobs justice and our nation will be
all the better for having him congratulations''}. However, when it comes
to the use of sentiments, differences between democratic and republican
voters can hardly be identified within the first sequence.
Pro-democratic as well as pro-republican voters vary regarding the use
of sentiments only within the confidence interval of each sequence (grey
area). Furthermore, the course of the slope, as well as the amplitude,
seem also to be influenced by other topics, as will be illustrated in
the second and third sequence.

\includegraphics{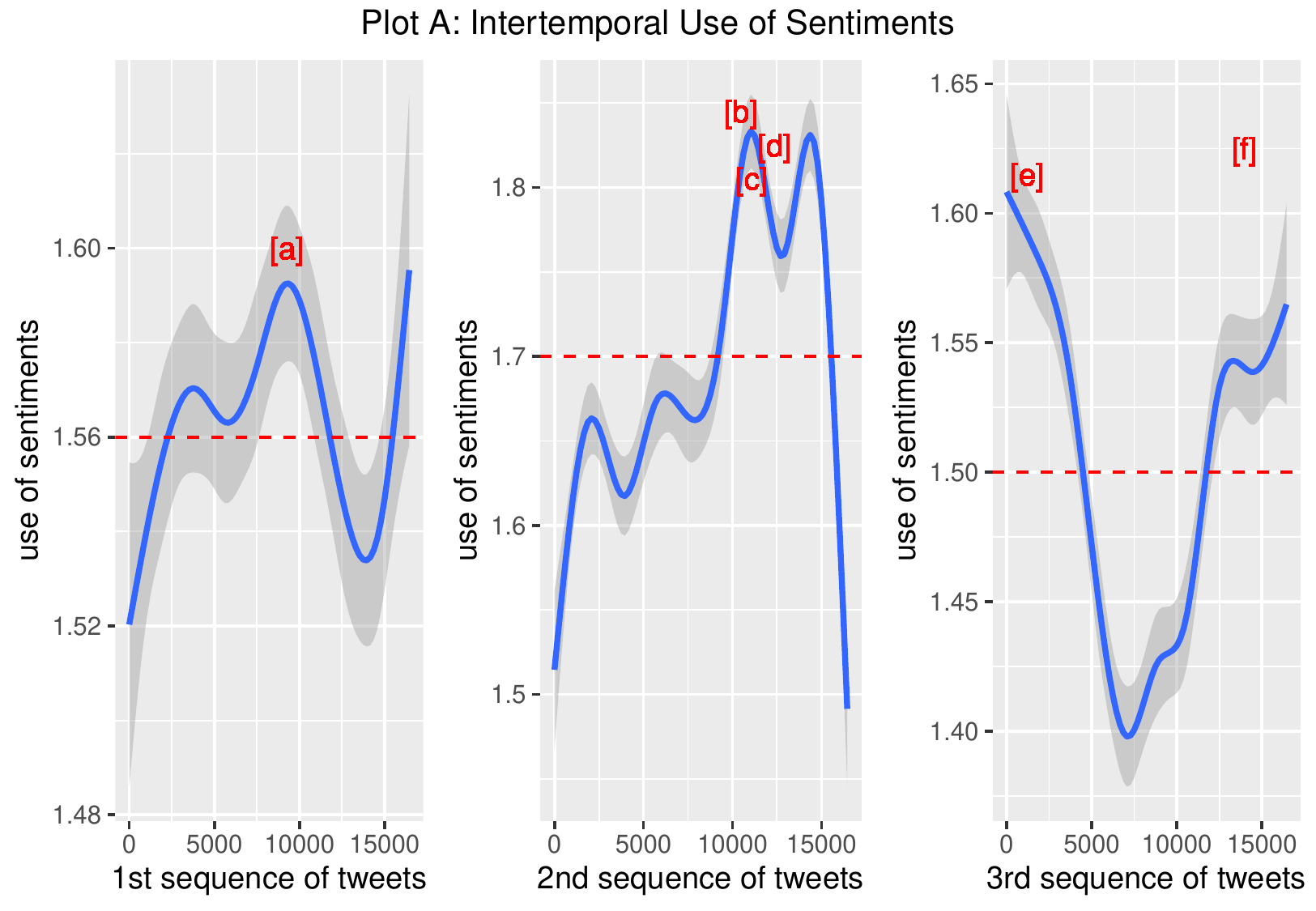}

As expected, the use of sentiments increases within the second sequence
and during the storming of the U.S. Capitol in Washington. As a result,
the amplitude increases and leads to two local maxima right after each
other. An intensive use of sentiments can be monitored, either to
support or to oppose this event. As a result, when it comes to
evaluating the use of sentiments from a numerical point of view,
supporters and opponents seem hardly to differ regarding their patterns
of communication. The following supportive Tweet is representative for
these developments ({[}b{]} Tweet 12.105): \emph{``we the people love
you mr president we admire your courage and determination to listen to
us and defend our rights thank you president trump for defending the
constitution respecting us and fighting for us you are our president all
50 states are red''}. Another Tweet represents the opponents ({[}c{]}
Tweet 12.183): \emph{``this isnt a peaceful protest this is an attack on
our democracy and domestic terrorism to try to stop certifying elections
this should be treated as a coup led by a president that will not be
peacefully removed from power''}. The majority of Tweets speaks out
against the storming of the U.S. Capitol in Washington and the use of
sentiments increases even more if people are directly affected by the
events ({[}d{]} Tweet 12.326): \emph{``i just had to evacuate my office
because of a pipe bomb reported outside supporters of the president are
trying to force their way into the capitol and i can hear what sounds
like multiple gunshots''}. Again, all Tweets from supporters and
opponents seem to vary within the confidence interval of the sequence.

The third sequence is characterized by many soothing and thus less
emotionally charged Tweets. As a result, the intertemporal use of
sentiments descends to the lowest values of the slope. Instead of
emotional outbursts, argumentative Tweets appear on both sides. At the
beginning of the third sequence some supporters of the storming of the
U.S. Capitol in Washington compose emotionally charged Tweets, but it is
important to note that they do not affect the course of the slope
({[}e{]} Tweet 778): \emph{``the corrupt democrats and educational
system efficiently spread their antiamerican cancer republicans stood
around like deer in the headlights you are the only one who fought it
total destruction of our democracy is ahead unless you continue
fighting''} Instead, the course of the slope is distorted by other
topics, such as sexual content ({[}f{]} Tweet 13.961): \emph{``seems my
sub is ready for a good dicking if i get 100 retweets maybe ill upload
my video of how i fucked his ass fuck interracial bdsm slave master''}.
The fact that such content has very different patterns of communication
than socio-political topics seems to shape a corresponding intertemporal
dynamic, which is illustrated by the amplitude. As a result, individual
Tweets as well as radical or extreme actors can be overshadowed on
Twitter and a case sensitive inspection seems required. The
intertemporal sentiment score can further clarify this.

\hypertarget{vii-intertemporal-sentiment-score}{%
\subsection{(VII) Intertemporal Sentiment
Score}\label{vii-intertemporal-sentiment-score}}

The same three sequences will be analyzed regarding the intertemporal
sentiment score. This is not about how many sentiments are used, but
whether they have a positive or negative connotation when it comes to
their sum within a Tweet. As a result, the first sequence shows that the
COVID-19 pandemic has more negative connotations than the outcome of the
elections in the federal state of Georgia. With a view to the first part
of the analysis, this means that fewer Tweets contain sentiments on the
subject of COVID-19, but their intertemporal score is quite negative.
Overall, the mean values within each sequence are negative, whereas the
first sequence contains the weakest negative mean value, because of the
outcome of the elections in Georgia. Again, the mean values do not cover
the dynamics of the sequences, as the amplitude of conditional means
suggests.

For example, supporters and opponents of a strict COVID-19 policy
slightly use negative sentiments, as can be demonstrated for the
supporters ({[}a{]} Tweet 1.717): \emph{``when do all the right wing
accounts and pundits paid by right wing billionaires apologize for
underplaying the covid virus''}. This can be shown for the opponents as
well ({[}b{]} Tweet 1.616): \emph{``the new lockdown is primarily about
giving the police leeway to be more draconian in enforcing it expect
many more examples of wanton authoritarianism and brutality to
follow''}. A moderate, but nevertheless slightly negative use of
language can be detected by focusing the sentiment score. With regard to
the outcome of the elections in Georgia, not only more sentiments can be
detected within Tweets, but these are also significantly more positive
overall, as could be shown in the previous step of the analysis. The low
confidence interval over the course of the slope within the first
sequence seems remarkable. This could mean that the patterns of
communication of supporters and opponents are similar regarding these
topics.

\includegraphics{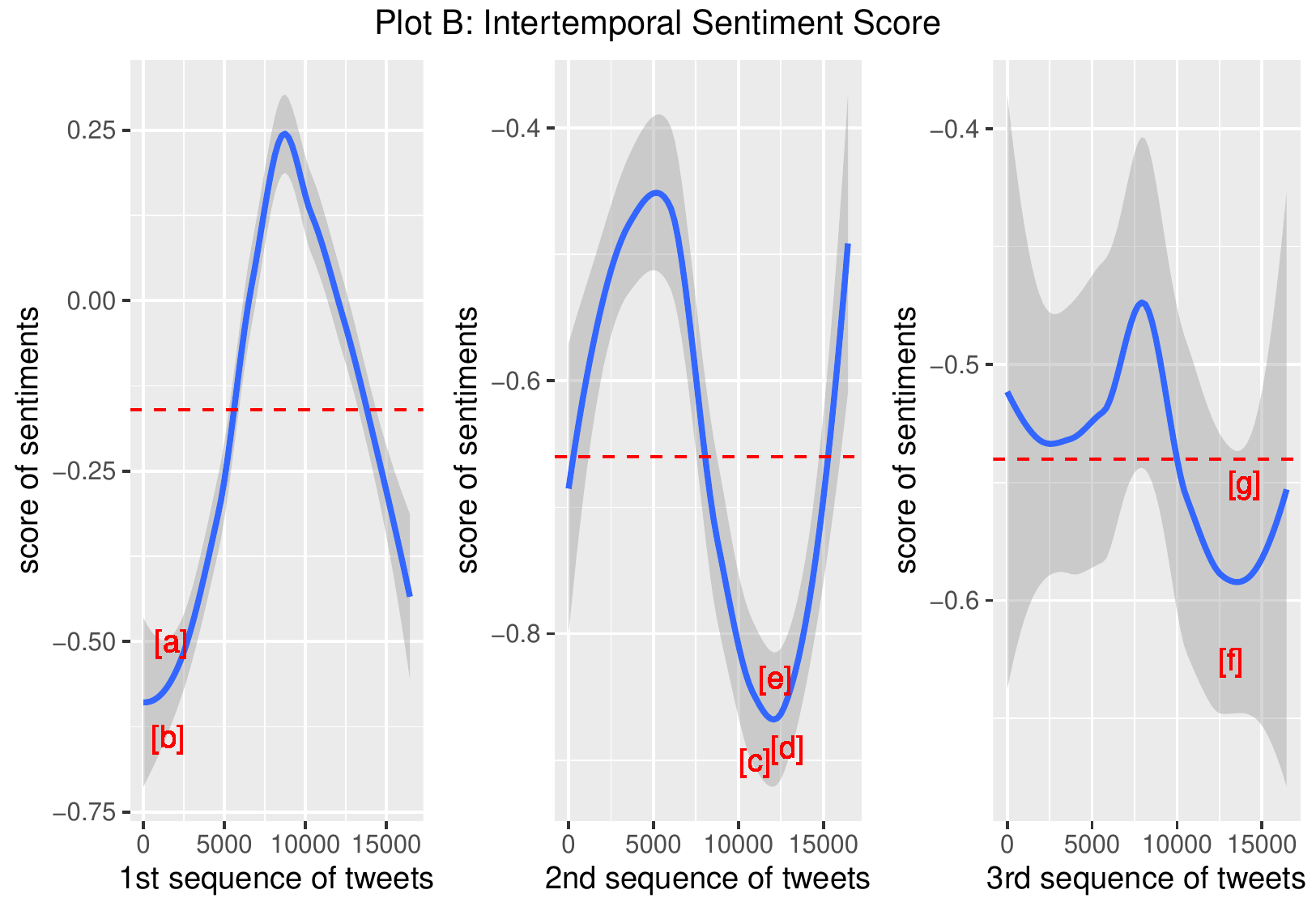}

The second sequence confirms that the storming of the U.S. Capitol in
Washington goes along with the local minimum of the slope regarding the
intertemporal sentiment score. Accordingly, the amplitude is strongest
in this sequence. The dominant emotional state at the local minimum of
this sequence is fear, as can be demonstrated with two examples of
opponents of the storming of the U.S. Capitol in Washington ({[}c{]}
Tweet 12.500): \emph{``i am in the capitol i am safe and my team and i
are sheltering in place the president of the united states has incited a
riot that has now stormed the capitol there are rioters roaming the
halls of the capitol i saw them with my own eyes our country deserves
better''} as well as ({[}d{]} Tweet 12.770) \emph{``i am safe we are
sheltering in place make no mistake president trump and his enablers are
directly responsible for this violence''}. Accordingly, the lowest
values of the slope are not to be detected after the storming of the
U.S. Capitol in Washington, but almost simultaneously. Again, the
intertemporal sentiment score shows a parallel between opponents and
supporters when it comes to the intensity of shown sentiments and the
underlying emotional states as well as their variance around the slope.
For example, supporters also refer to fear and trust as emotional state
and use it within their Tweets at a comparable rate ({[}e{]} Tweet
12.689): \emph{``no matter what happens today stay strong do not waiver
do not walk in fear no matter what it looks like hold the line of faith
trump is our president''}. However, there is no difference between the
Tweets of the so-called Trumpists and other supporters of the event.
Their patterns of communication vary only in accordance with the course
of the slope.

One of the most significant dynamics is revealed in the third sequence.
The confidence interval around the slope seems to increase
significantly. This does not primarily indicate that supporters and
opponents after the storming of the U.S. Capitol in Washington tend to
have more differences regarding their intertemporal sentiment score, but
can rather be interpreted as a signal for more emerging topics within
the Tweets. As illustrated in the previous step of the analysis, the
intertemporal use of sentiments decreases in this sequence. Therefore,
the amplitude has decreased as well compared to the previous sequence.
This is another indication that sensitivity is required for
corresponding analyses. On the one hand, Twitter users seem to make use
of fewer sentiments, on the other hand, they seem to differ more clearly
regarding the strength of the sentiments expressed. This can be
interpreted as a slight after-effect in respect to the current events.
Furthermore, the length of the Tweets decrease and the topics return to
the topics of the first sequence, e.g.~the necessity of measurements
regarding the COVID-19 pandemic in combination with the storming of the
U.S. Capitol in Washington ({[}f{]} Tweet 13.987): \emph{``not a mask to
be seen at the trump sore loser rally''}. The contents of the Tweets
also seem to be devoted more to a substantive debate ({[}g{]} Tweet
14.069): \emph{``if masks are so great why do the mask mandates show
growth in the graphs after implementation''}.

As a result, and in accordance with the first part of the analysis,
events deviating from the current events with more sentiments and
emotional states can lead to a significant distortion of the slope. This
makes it harder to spot promoters of events like the storming of the
U.S. Capitol in Washington. Likewise, it seems harder to differentiate
between supporters and opponents by relying solely on quantitative
indicators.

\hypertarget{viii-summary}{%
\subsection{(VIII) Summary}\label{viii-summary}}

The analysis has shown that differences between supporters and opponents
of relevant social affairs, like the storming of the U.S. Capitol in
Washington, can be similar in their patterns of communication. In fact,
they can even be so similar that it seems almost impossible to
differentiate them solely based upon quantitative characteristics. This
similarity was determined via the number of sentiments and emotional
states used over time, but also via their associated summative score. As
a result, the values of individual Tweets vary most likely within the
conditional mean values and conditional variance values, even if their
content supports such events. Therefore, all Tweets from supporters and
opponents vary within the confidence interval of each sequence. This is
due to the fact that the intertemporal dynamics are affected by social
affairs and corresponding Tweets vice versa. The assumption that
extremist Tweets or hate speech can be identified by above-average
quantitative values would therefore be wrong. Furthermore it would be
wrong to use mean values and variance values without considering the
intertemporal dynamics framed by the context. This could result in
false-positive identifications in the context of radicalization
research.

Furthermore, the analysis demonstrated that intertemporal dynamics
influence the course as well as the amplitute of the slope. This
influence is not exclusively due to social affairs or similar events.
Topics with different patterns of communication, like sexual content,
can significantly influence the intertemporal dynamics as well.
Especially the permeability of social media platforms like Twitter and
the interaction between different echo chambers could not only affect
the course of the slope globally (Cinelli et al.~2021), but also the
intertemporal dynamics partially within the echo chambers. Thereby,
relevant phenomena for the context of radicalization research can be
overshadowed.

This confirms the two hypotheses that intertemporal dynamics can be
traced and obscure relevant phenomena in the context of radicalization
research. As a result, longitudinal sentiment analyses seem less
suitable for the targeted identification of individual Tweets, but more
suitable for depicting a development over time in the sum of all Tweets.
This is in accordance with the findings of Grogan (2020) as well as the
suggestion made by Greipl et al.~(2022) to conduct longitudinal analyses
in radicalization research. The developments can be mapped almost in
real time, which offers the possibility for qualitative inspections of
Tweets, which seems necessary. Accordingly, the importance of
qualitative perspectives was appropriately emphasized in the anthology
of Hamachers et al.~(2020). Finally, the question arises, whether the
similarities found between the supporters and opponents of the storming
of the U.S. Capitol in Washington are not a result of the predefined
structures of social media platforms. As a social media platform,
Twitter specifies the same input format for all its users, whereby their
patterns of communication can be influenced as well.

\hypertarget{ix-recommendations}{%
\subsection{(IX) Recommendations}\label{ix-recommendations}}

Taking into account the permeability of echo chambers on social media
platforms and their intertemporal dynamics, a longitudinal approach
seems necessary in order to depict process-based phenomena in the
context of radicalization research. With regard to sentiment analysis
and in accordance with current research, it can be stated that this
methodological approach seems not suitable for a precise identification
of these phenomena outside their echo chambers, either in a
cross-sectional nor in a longitudinal perspective. However, sentiment
analyses seem rather suitable for depicting the intertemporal dynamics
on social media platforms in general. By doing so, the use of several
sentiment analysis dictionaries in order to validate the findings has
shown to be a beneficial factor.

It could also be shown that phenomena relevant for the context of
radicalization research do not necessarily have the strongest influence
on the intertemporal dynamics, but other phenomena could rather
superimpose them. Therefore, it would be negligent to identify
radicalization, extremism and hate speech solely on the basis of
above-average quantitative values. The distribution of these phenomena
seems rather to depend on the fact that they are within the range of
conditional mean values and conditional variance values. One could also
say that these users seem to have mastered the rules of social media
platforms. As a result, in this example, the Tweets from Trumpists,
Republicans and Democrats are quite similar regarding the storming of
the U.S. Capitol in Washington. If a social event elicits increased
activity from one side, it appears to do the same for the other side.
Accordingly, the highs and lows of these intertemporal dynamics should
be a reminder for a deeper insight via qualitative research and human
expertise.

\hypertarget{sources}{%
\subsection{Sources}\label{sources}}

\begin{itemize}
\item
  Auguie, Baptise \& Anton Antonov (2017): ``gridExtra: Miscellaneous
  Functions for''Grid'' Graphics''. Online:
  \url{https://cran.r-project.org/package=gridExtra}
\item
  Awan, Imran (2017): ``Cyber-Extremism: Isis and the Power of Social
  Media''. Society, 54(3). Online:
  \url{https://link.springer.com/article/10.1007/s12115-017-0114-0}
\item
  Bache, Stefan Milton \& Hadley Wickham (2020): ``magrittr: A
  Forward-Pipe Operator for R''. Online:
  \url{https://cran.r-project.org/package=magrittr}
\item
  Barberá, Pablo; Jost, John; Nagler, Jonathan; Tucker, Joshua; \&
  Richard Bonneau (2015): ``Tweeting From Left to Right: Is Online
  Political Communication More Than an Echo Chamber?'' Psychological
  Science, 26(10). Online:
  \url{https://doi.org/10.1177/0956797615594620}
\item
  Bretschneider, Uwe \& Ralf Peters (2017): ``Detecting Offensive
  Statements towards Foreigners in Social Media''. International
  Conference on System Sciences:
  \url{http://dx.doi.org/10.24251/HICSS.2017.268}
\item
  Bright, Jonathan (2017): ``Explaining the emergence of echo chambers
  on social media: the role of ideology and extremism''. Online:
  \url{https://arxiv.org/abs/1609.05003}
\item
  Charitidis, Polychronis; Doropoulos, Stavros; Vologiannidis, Stavros;
  Papastergiou, Ioannis, \& Sophia Karakeva (2020): ``Towards countering
  hate speech against journalists on social media''. Online Social
  Networks and Media, 17. Online: \url{https://arxiv.org/abs/1912.04106}
\item
  Cinelli, Matteo; Morales, Gianmarco De Francisci; Galeazzi,
  Alessandro; Quattrociocchi, Walter \& Michele Starnini (2021): ``The
  echo chamber effect on social media''. Online:
  \url{https://doi.org/10.1073/pnas.2023301118}
\item
  De Queiroz, Gabriela; Fay, Colin; Hvitfeldt, Emil; Keyes, Os; Misra,
  Kanishka; Mastny, Tim; Erickson, Jeff; Robinson, David \& Julia Silge
  (2022): ``tidytext: Text Mining using''dplyr'', ``ggplot2'', and Other
  Tidy Tools''. Online:
  \url{https://cran.r-project.org/package=tidytext}
\item
  Feller, William (1991): ``An Introduction to Probability Theory and
  its Applications''. 2nd Edition. Wiley Series in Probability and
  Statistics. ISBN: 978-0-471-25709-7
\item
  Gerstenfeld, Phyllis; Grant, Diana \& Chau-Pu Chiang (2003): ``Hate
  Online: A Content Analysis of Extremist Internet Sites''. Analyses of
  Social Issues and Public Policy, 1. Online:
  \url{https://doi.org/10.1111/j.1530-2415.2003.00013.x}
\item
  Glaser, Stefan \& Thomas Pfeiffer (2017): ``Erlebniswelt
  Rechtsextremismus: modern - subversiv - hasserfüllt. Hintergründe und
  Methoden für die Praxis der Prävention''. 5. Auflage. Wochenschau
  Verlag. ISBN: 978-3-7344-0500-6
\item
  Greipel, Simon; Hohner, Julian; Schulze, Heidi \& Diana Rieger (2022):
  Radikalisierung im Internet: Ansätze zur Differenzierung, empirische
  Befunde und Perspektiven zu Online-Gruppendynamiken. In: MOTRA-Monitor
  2021. ISBN: 978-3-9818469-4-2
\item
  Grogan, Michael (2020): ``NLP from a time series perspective. How time
  series analysis can complement NLP''. Towards Data Science. Online:
  \url{https://towardsdatascience.com/nlp-from-a-time-series-perspective-39c37bc18156}
\item
  Hamachers, Annika; Weber, Kristin \& Stephan Jarolimek (2020):
  ``Extremistische Dynamiken im Social Web''. Verlag für
  Polizeiwissenschaft. ISBN: 978-3-86676-671-6
\item
  Hamborg, Felix \& Karsten Donnay (2021): ``NewsMTSC: A Dataset for
  (Multi-)Target-dependent Sentiment Classification in Political News
  Articles''. Proceedings of the 16th Conference of the European Chapter
  of the Association for Computational Linguistics. Online:
  \url{https://aclanthology.org/2021.eacl-main.142/}
\item
  Hu, Minqing / Bing Liu (2004): ``Mining and Summarizing Customer
  Reviews''. University of Illinois at Chicago (Academic Press). Online:
  \url{https://www.cs.uic.edu/~liub/publications/kdd04-revSummary.pdf}
\item
  Jacobs, Kristof \& Niels Spierings (2018): ``A populist paradise?
  Examining populists' Twitter adoption and use''. Information,
  Communication \& Society, 22(12). Online:
  \url{https://doi.org/10.1080/1369118X.2018.1449883}
\item
  Kay, Jonathan (2011): ``Among the Truthers: A Journey Through
  America's Growing Conspiracist Underground''. New York: HarperCollins.
  ISBN: 978-0062004819
\item
  Klinkhammer, Dennis (2020): ``Analysing Social Media Network Data with
  R: Semi-Automated Screening of Users, Comments and Communication
  Patterns''. Online: \url{https://arxiv.org/abs/2011.13327}
\item
  Klinkhammer, Dennis (2022): ``Sentiment Analysis with R: Natural
  Language Procesing for Semi-Automated Assessments of Qualitative
  Data''. Online: \url{https://arxiv.org/abs/2206.12649}
\item
  Korkontzelos, Ioannis; Nikfarjam, Azadeh; Shardlow, Matthew; Sarker,
  Abeed; Ananiadou, Sophia \& Graciela Gonzalez (2016): ``Analysis of
  the effect of sentiment analysis on extracting adverse drug reactions
  from tweets and forum posts''. Journal of Biomedical Informatics, 62.
  Online: \url{https://doi.org/10.1016/j.jbi.2016.06.007}
\item
  Machackova, Hana; Blaya, Catherine; Bedrosova, Marie; Smahel, David \&
  Elisabeth Staksrud (2020): ``Children's experiences with cyberhate''.
  Online:
  \url{https://www.lse.ac.uk/media-and-communications/assets/documents/research/eu-kids-online/reports/euko-cyberhate-22-4-final.pdf}
\item
  Mandl, Thomas; Modha, Sandip; Majumder, Prasenjut; Patel, Daksh; Dave,
  Mohana; Mandlia, Chintak \& Aditya Patel (2019): ``Overview of the
  HASOC track at FIRE 2019: Hate Speech and Offensive Content
  Identification in Indo-European Languages''. 11th Forum for
  Information Retrieval Evaluation. Online:
  \url{https://doi.org/10.1145/3368567.3368584}
\item
  Mohammad, Saif M. (2020): ``Sentiment Analysis: Automatically
  Detecting Valence, Emotions, and Other Affectual States from Text''.
  Online: \url{https://arxiv.org/abs/2005.11882}
\item
  Neumann, Peter (2013): ``The Trouble with Radicalization''.
  International Affairs, 89(4). Online:
  \url{https://doi.org/10.1111/1468-2346.12049}
\item
  Neumann, Peter; Winter, Charlie; Meleagrou-Hitchens, Alexander;
  Ranstorp, Magnus \& Lorenzo Vidino (2018): ``Die Rolle des Internets
  und sozialer Medien für Radikalisierung und Deradikalisierung''. PRIF
  Report, 9. ISBN: 978-3946459385
\item
  O'Hara, Kieron, \& David Stevens (2015): ``Echo Chambers and Online
  Radicalism: Assessing the Internet's Complicity in Violent
  Extremism''. Policy \& Internet, 7(4). Online:
  \url{https://doi.org/10.1002/poi3.88}
\item
  Ogneva, Maria (2010): ``How Companies Can Use Sentiment Analysis to
  Improve Their Business''. Online:
  \url{https://mashable.com/archive/sentiment-analysis}
\item
  Pereira-Kohatsu, Juan Carlos, Quijano-Sánchez, Lara, Liberatore,
  Federico \& Miguel Camacho-Collados (2019): ``Detecting and Monitoring
  Hate Speech in Twitter''. Sensors, 19(21). Online:
  \url{https://doi.org/10.3390/s19214654}
\item
  Prior, Markus (2005): ``News vs.~Entertainment: How Increasing Media
  Choice Widens Gaps in Political Knowledge and Turnout''. American
  Journal of Political Science, 49(3). Online:
  \url{https://doi.org/10.2307/3647733}
\item
  Reichelmann, Ashley; Hawdon, James; Costello, Matt; Ryan, John; Blaya,
  Catherine; Llorent, Vincente; Oksanen, Atte; Räsänen, Pekka \& Izabela
  Zych (2020): ``Hate Knows No Boundaries: Online Hate in Six Nations''.
  Online: \url{https://doi.org/10.1080/01639625.2020.1722337}
\item
  Ross, Björn; Rist, Michael; Carbonell, Guillermo; Cabrera, Ben;
  Kurowsky, Nils \& Michael Wojatzki (2017): ``Measuring the Reliability
  of Hate Speech Annotations: The Case of the European Refugee Crisis''.
  Duisburg-Essen: University of Duisburg-Essen. Online:
  \url{https://arxiv.org/abs/1701.08118}
\item
  Scanlon, Jacob \& Matthew Steven Gerber (2015): ``Forecasting violent
  extremist cyber recruitment''. IEEE Transactions on Information
  Forensics and Security, 10(11). Online:
  \url{http://dx.doi.org/10.1109/TIFS.2015.2464775}
\item
  Schmidt, Anna, \& Wiegand, Michael (2017): ``A Survey on Hate Speech
  Detection using Natural Language Processing''. 5th International
  Workshop on Natural Language Processing for Social Media. Online:
  \url{http://dx.doi.org/10.18653/v1/W17-1101}
\item
  Sui, Xueqin; Chen, Zhumuin; Wu, Kai; Ren, Pengjie; Ma, Jun \& Fenyu
  Zhou (2014): ``Social media as sensor in real world: Geolocate user
  with microblog''. Communications in Computer and Information Science,
  496. Online: \url{http://dx.doi.org/10.1007/978-3-662-45924-9_21}
\item
  Sunstein, Cass (2006): ``Infotopia: How Many Minds Produce
  Knowledge''. Oxford: Oxford University Press. ISBN: 978-0195340679
\item
  Tanoli, Irfan; Pais, Sebastiao; Cordeiro, Joao \& Muhammad Luqman
  Jamil (2022): ``Detection of Radicalisation and Extremism Online: A
  Survey''. Online:
  \url{https://assets.researchsquare.com/files/rs-1185415/v1_covered.pdf}
\item
  Torregrosa, Javier; Bello-Orgaz, Gema; Martínez-Cámara, Eugenio; Del
  Ser, Javier \& David Camacho (2022): ``A survey on extremism analysis
  using natural language processing: definitions, literature review,
  trends and challenges''. Online:
  \url{https://link.springer.com/article/10.1007/s12652-021-03658-z}
\item
  Tumasjan, Andranik; Sprenger, Timm O.; Sandner, Philipp G. \& Isabelle
  M. Welpe (2010): ``Predicting Elections with Twitter: What 140
  Characters Reveal about Political Sentiment''. Proceedings of the
  Fourth International AAAI Conference on Weblogs and Social Media.
  ISBN: 978-1-57735-445-1
\item
  Warner, William \& Julia Hirschberg (2012): ``Detecting Hate Speech on
  the World Wide Web''. Proceedings of the Second Workshop on Language
  in Social Media. Online: \url{https://aclanthology.org/W12-2103/}
\item
  Wickham, Hadley (2019): ``stringr: Simple, Consistent Wrappers for
  Common String Operations''. Online:
  \url{https://cran.r-project.org/package=stringr}
\item
  Wickham, Hadley; François, Roman \& Kirill Müller (2022): ``dplyr: A
  Grammmar of Data Manipulation''. Online:
  \url{https://cran.r-project.org/package=dplyr}
\item
  Wickham, Hadley; Chang, Winston; Henry, Lionel; Lin Pedersen, Thomas;
  Takahashi, Kohske; Wilke, Claus; Woo, Kara; Yutani, Hiroaki \& Dewey
  Dunnington (2022): ``ggplot2: Create Elegant Data Visualisations Using
  the Grammar of Graphics''. Online:
  \url{https://cran.r-project.org/package=ggplot2}
\item
  Wiegand, Miachael; Siegel, Melanie \& Josef Ruppenhofer (2018):
  ``Overview of the GermEval 2018 Shared Task on the Identification of
  Offensive Language''. Saarbrücken: University of Saarland Press.
  Online:
  \url{https://www.lsv.uni-saarland.de/wp-content/publications/2018/germeval2018_wiegand.pdf}
\item
  Wienigk, Ruben \& Dennis Klinkhammer (2021): ``Online-Aktivitäten der
  Identitären Bewegung auf Twitter - Warum Kontensperrungen die Anzahl
  an Hassnachrichten nicht reduzieren''. Forum Kriminalprävention.
  Online:
  \url{https://www.forum-kriminalpraevention.de/online-aktivitaeten-der-identitaeren-bewegung.html}
\item
  Wood, Ian B.; Varela, Pedro Leal; Bollen, Johan; Rocha, Luis M. \&
  Joana Gonçalves-Sá (2017): ``Human Sexual Cycles are Driven by Culture
  and Match Collective Moods''. Online:
  \url{https://arxiv.org/abs/1707.03959}
\end{itemize}

\hypertarget{affiliations}{%
\paragraph{Affiliations}\label{affiliations}}

Dennis Klinkhammer is Professor for Empirical Research at the FOM
University of Applied Sciences. He advises public as well as
governmental organisations on the application of multivariate statistics
and limitations of artificial intelligence by providing introductions to
Python and R: \url{https://www.statistical-thinking.de} (Homepage)
\url{https://github.com/statistical-thinking} (GitHub)

\end{document}